\pdfoutput=1

\documentclass[11pt]{article}

\usepackage[]{EMNLP2022}

\usepackage{times}
\usepackage{latexsym}
\usepackage{comment}
\usepackage{pgf-pie} 
\usepackage{caption}
\usepackage{amsmath}
\usepackage{makecell}
\usepackage{xcolor}

\usepackage[T1]{fontenc}

\usepackage[utf8]{inputenc}

\usepackage{microtype}

\usepackage{inconsolata}

\usepackage{booktabs} 
\usepackage{tabularx} 
\usepackage{multirow}
\usepackage{makecell}
\usepackage{graphicx}
\usepackage{makecell}

\title{Detecting Unintended Social Bias in Toxic Language Datasets}

\author{Nihar Sahoo$^{\ast}$, Himanshu Gupta$^{\ast}$, Pushpak Bhattacharyya \\
CFILT, Indian Institute of Technology Bombay, India\\
\texttt{\{nihar, himanshug, pb }@cse.iitb.ac.in\}\\ }

\begin{document}
\maketitle
\def\thefootnote{*}\footnotetext{These authors contributed equally to this work}\def\thefootnote{\arabic{footnote}}
\begin{abstract}
\textcolor{red}{\textit{\textbf{Warning:} This paper has contents which may be offensive, or upsetting however this cannot be avoided owing to the nature of the work. }}

With the rise of online hate speech, automatic detection of Hate Speech, Offensive texts as a natural language processing task is getting popular. However, very little research has been done to detect unintended social bias from these toxic language datasets. This paper introduces a new dataset \textit{ToxicBias} curated from the existing dataset of Kaggle competition named "Jigsaw Unintended Bias in Toxicity Classification". We aim to detect social biases, their categories, and targeted groups. The dataset contains instances annotated for five different bias categories, viz., \textit{gender, race/ethnicity, religion, political, and LGBTQ}. We train transformer-based models using our curated datasets and report baseline performance for bias identification, target generation, and bias implications. Model biases and their mitigation are also discussed in detail. Our study motivates a systematic extraction of social bias data from toxic language datasets. All the codes and dataset used for experiments in this
work are publicly available\footnote{\url{https://github.com/sahoonihar/ToxicBias_CoNLL_2022}}.
\end{abstract}

\section{Introduction} 
In the age of social media and communications, it is simpler than ever to openly express one's opinions on a wide range of issues. This openness results in a flood of useful information that can assist people in being more productive and making better decisions. According to statista  \footnote{\url{https://www.statista.com/statistics/278414/number-of-worldwide-social-network-users/}}, the global number of active social media users has just surpassed four billion, accounting for more than half of the world’s population. The user base is expected to grow steadily over the next five years. Various studies \cite{teen-impact} say that children and teenagers, who are susceptible, make up a big share of social media users. Unfortunately, this increasing number of social media users also leads to an increase in toxicity \cite{social-media}. Sometimes this toxicity gives birth to violence and hate crimes. It does not just harm an individual; most of the time, the entire community suffers as due to its intensity.
\begin{figure}
    \centering
    \includegraphics[width = \columnwidth]{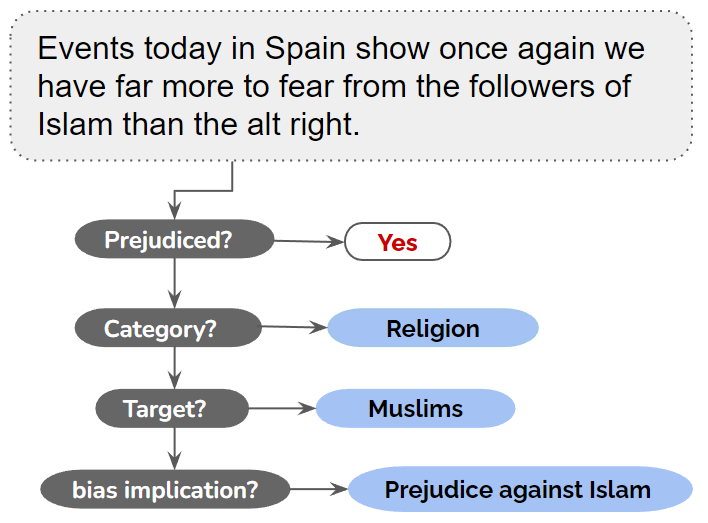}
    \caption{An illustrative example of \textit{ToxicBias}. During the annotation process, hate speech/offensive text is provided without context. Annotators are asked to mark it as biased/neutral and to provide category, target, and implication if it has biases. }
    \label{fig:Dataset Example}
\end{figure}

We have different perspectives based on race, gender, religion, sexual orientation, and many other factors. These perspectives sometimes lead to biases that influence how we see the world, even if we are unaware of them. Biases like this can lead us to make decisions that are neither intelligent nor just. Furthermore, when these biases are expressed as hate speech and offensive texts, it becomes painful for specific communities. While some of these biases are implied, most explicit biases can be found in the form of hate speech and offensive texts. 

The use of hate speech incites violence and sometimes leads to societal and political instability. BLM (Black Lives Matter) movement is the consequence of one such bias in America. So, to address these biases, we must first identify them. While the concepts of Social Bias and Hate Speech may appear to be the same, there are subtle differences.

This paper expands on the above ideas and proposes a new dataset \textbf{\textit{ToxicBias}} for detecting social bias from toxic language datasets. The main contributions can be summarized as follows:

\begin{itemize}
    \item To the best of our knowledge, this is the first study to extract social biases from toxic language datasets in English.
    \item We release a curated dataset of 5409 instances for detection of social bias, its categories, targets and bias reasoning.
    \item We present methods to reduce lexical overfitting using counter-narrative data augmentation.
\end{itemize}

In the following section we discuss various established works which are aligned with our work. Section \ref{sec:dataset} provides information about our dataset, terminology, annotation procedure, and challenges. In section \ref{sec:dataset}, we describe our tests and results, followed by a discussion of lexical overfitting reduction via data augmentation in section \ref{sec:model_bias}. Section \ref{sec:con} discusses the conclusion and future works.
\section{Related Work}

\noindent \textbf{Offensive Text:} Unfortunately, offensive content poses some unique challenges
to researchers and practitioners. First and foremost, determining what constitutes abuse/offensive behaviour is difficult. Unlike other types of malicious activity, e.g., spam or malware, the accounts carrying out this type of behavior are usually controlled by humans, not bots \cite{founta2018large}.The term “offensive language” refers to a broad range of content, including hate speech, vulgarity, threats, cyberbully, and other ethnic and racial insults \cite{KAUR2021274}. There is no single definition of abuse, and phrases like "harassment," "abusive language," and "damaging speech" are frequently used interchangeably.

\noindent \textbf{Hate Speech:} Hate Speech is defined as speech that targets disadvantaged social groups in a way that may be damaging to them. \cite{davidson2017automated}. \citet{article} defines Hate speech as follows: "Hate speech is a language that attacks or diminishes, that incites violence or hate against groups, based on specific characteristics such as physical appearance, religion, national or ethnic origin, sexual orientation, gender identity or other, and it can occur with different linguistic styles, even in subtle forms or when humor is used".

\noindent \textbf{Bias in Embedding:} The initial works to explore bias in language representations aimed at detecting gender, race, religion biases in word representations \cite{Bolukbasi,caliskan,manzini-etal}. Some of recent works have focused on bias detection from sentence representations \cite{may-etal,kurita2019measuring} using BERT embedding. 

In addition, there have been a lot of notable efforts towards detection of data bias in hate speech and offensive languages \cite{waseem-hovy,davidson-19,sap-etal-2019-risk,Mozafari}. \citet{borkan} has discussed the presence of unintended bias in hate speech detection models for identity terms like islam, lesbian, bisexual, etc. The biased association of different marginalized groups is still a major challenge in the models trained for toxic language detection \cite{kim2020intersectional,xia-etal-2020-demoting}. This is mainly due to the bias in annotated data which creates the wrong associations of many lexical features with specific labels \cite{dixon}. Lack of social context of the post creator also affect the annotation process leading to bias against certain communities in the dataset \cite{sap-etal-2019-risk}.

\noindent \textbf{Social bias datasets:} More recently, many datasets \cite{nadeem-etal-2021-stereoset,nangia2020crows} have been created to measure and detect social biases like gender, race, profession, religion, age, etc. However, \citet{blodgett2021stereotyping} has reported that many of these datasets lack clear definitions and have ambiguities and inconsistencies in annotations. A similar study have been done in \cite{sap2020social}, where dataset has both categorical and free-text annotation and generation framework as core model.

There have been few studies on data augmentation \cite{nozza,bartl} to decrease the incorrect association of lexical characteristics in these datasets. \citet{toxigen} proposed a prompt based framework to generate large dataset of toxic and neutral statements to reduce the spurious correlation for Hate Speech detection.

However, no study has been done for detecting social biases from toxic languages, which is a challenging task due to the conceptual overlap between hate speech and social bias. Using a thorough guideline, we attempt to uncover harmful biases in toxic language datasets. The curated dataset is discussed in length in the next section, as are the definitions of each category label and the annotation procedure.

\section{ToxicBias Dataset} \label{sec:dataset}
We develop the manually annotated \textit{ToxicBias} dataset to enable the algorithm to correctly identify social biases from a publicly available toxicity dataset. Below, we define social bias and the categories taken into account in our dataset. The comprehensive annotation process that we use for dataset acquisition is then covered.
\subsection{Social Bias}
People typically have preconceptions, stereotypes, and discrimination against other who do not belong to their social group. Positive and negative social bias refers to a preference for or against persons or groups based on their social identities (e.g., race, gender, etc.). Only the negative biases, however, have the capacity to harm target groups \cite{trouble-with-bias}. As a result, in our study, we \textit{focus on identifying negative biases} in order to prevent harmful repercussions on targeted groups. Members of specific social groups (e.g., Women, Muslims, and Transgender individuals) are more likely to face prejudice as a result of living in a culture that does not sufficiently support fairness. In this work, we have considered five prevalent social biases:
\begin{itemize}
\itemsep0em 
    \item\textbf{Gender:} Favoritism towards one gender over other. It can be of the following types: Alpha, Beta or Sexism \cite{park-etal-2018-reducing}.
    \item\textbf{Religion:} Bias against individuals on the basis of religion or religious belief. e.g. Christianity, Islam, Scientology etc \cite{religion-bias}.

    \item\textbf{Race:} Favouritism for a group of people having  common visible physical traits, common origins, language etc. It is related to  dialect, color, appearance, regional or societal perception \cite{sap-etal-2019-risk}.  

    \item\textbf{LGBTQ:}  Prejudice towards LGBTQ community people. It can be due to societal perception or physical appearance. 

    \item\textbf{Political:} Prejudice against/towards individuals on the basis of their political beliefs. For example: liberals, conservatives, etc.
\end{itemize}
For all of these categories, target terms are the communities towards which bias is targeted.

\begin{table}
\centering
\begin{tabular}{ll}
\hline
Categories & Targets\\ \hline
Political & liberal, conservative, feminist, etc. \\
Religion & christian, jew, hindu, atheist, etc.  \\
Gender & men, women \\
LGBTQ & gay, lesbian, homosexual, etc. \\
Race & black, white, asian, canadians, etc.\\ \hline
\end{tabular}
    \caption{Bias categories and corresponding targets.}
    \label{tab:Annotation categories and targets}
\end{table}

\subsection{Social Bias \textit{Vs} Hate Speech}
While Social Bias and Hate Speech may appear the same at first look, they are not. The differences between them are quite subtle. While hate speech is always associated with negative sentiment, social bias can also have positive sentiments. Social bias is preconceived belief toward or against specific social identities, whereas hate speech is an explicit comment expressing hatred against an individual or a group. Not all hate speech is biased, and not all biased speech is hate speech. We will use the following examples to demonstrate the differences:
\begin{itemize}
\itemsep0em 
    \item Some comments are merely toxic without containing any social biases in them, e.g. \\
    \textit{IM FREEEEE!!!! WORST EXPERIENCE OF MY F**K-ING LIFE}
    \item Toxic comments can be hate speech but not necessarily biased, e.g.\\
    \textit{you gotta be kidding.  trump a Christian, nope, he is the devil, he hates blacks, Hispanics, muslims, gays, Asians, etc.}
    \item Some comments are just biased with negating sentiment without containing any toxicity or hate speech in them, e.g. \\
    \textit{All Asian people are bad drivers.}
\end{itemize}

\subsection{Annotation Process}
\begin{table*}
\resizebox{\textwidth}{!}{%
\centering
\begin{tabular}{lllll}
\hline
        comment\_text & label & category & target & implication\\ \hline
        I wish gay men would just leave children alone. & bias & lgbtq & gays & gays are pedophiles\\
        Islam, unfortunately, is a violent cult.& bias & religion & muslims & muslims are violent\\
        Why does Netflix hate gay people? & neutral &-&-&-\\
        Liberal women are so \#\^king ugly! & bias & political, gender & liberals, women & prejudice against liberal women\\
        \makecell[l]{Why are black thugs coming out here murdering locals? What\\ is the real story?  Please someone..hire an investigative journalist.} & bias & race & blacks & prejudice against blacks\\
\hline
\end{tabular}}
\caption{\label{Bias Data Annotation}Annotation of Bias along with its category, targeted groups and implication}
\end{table*}
The dataset we used for annotation is collected from a Kaggle competition named "Jigsaw Unintended Bias in Toxicity Classification \cite{jigsaw, research}". It has around two million Wikipedia comments annotated for toxicity. The data also has several other toxicity subtype attributes such as \textit{severe toxicity, obscene, threat, insult, identity\_attack, and sexual\_explicit}.

We discovered that, with the exception of the \textit{identity\_attack} column, all of the columns in this dataset are redundant for the social bias detection task. \textit{Identity\_attack} shows denigration of the comment towards a identity. This identity could be related to race, religion, political affiliation, gender, or sexual orientation. So, we filtered out only those comments for which \textit{identity\_attack} values are greater than or equal to $0.5$. We annotated this filtered dataset for the presence of social bias. We have considered only \textit{five bias categories} for our annotation and \textit{possible targets} listed in Table \ref{tab:Annotation categories and targets}. We did not include other categories due to their low presence in the original dataset. The targets describe any social or demographic groups that is targeted in the comment. Bias implications are annotated in addition to bias categories and relevant targets. Table \ref{Bias Data Annotation} shows a sample annotation of this filtered dataset. The bias implications are simple \textit{free-text} reasons showing the stereotype towards the target group.

\begin{figure}
    \centering
    \includegraphics[width = \columnwidth]{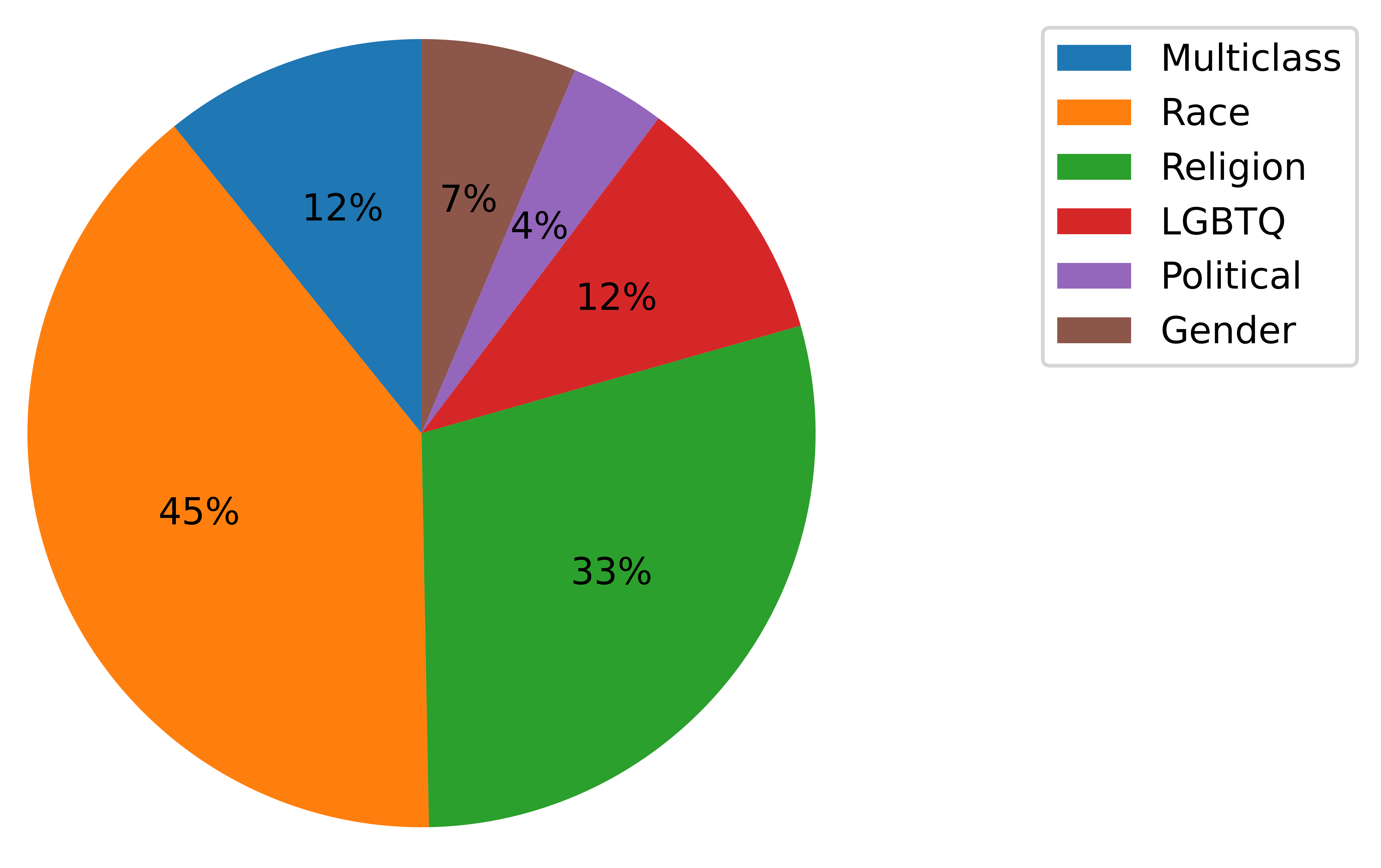}
    \caption{Distribution of bias categories in ToxicBias. It is observed that some instances qualified for multiple bias categories$(12.22\%)$}
    \label{fig:Dataset dist}
\end{figure}

The final dataset contains 5409 cases with multiple label annotations. There are 120 distinct terms for target annotation divided into five categories. To check the consistency of our framework and to categorize biases, two different annotators annotated the data independently. Considering the complexity of the task, we provided a detailed guideline to each of the annotators. Following the thorough guidelines by \citet{singh-etal-2022-hollywood}, we developed a series of questionnaires for each categories to assist the annotators. Inter-annotator agreement was assessed for the first $2500$ occurrences, and a Cohen's Kappa value of $64.3$ was found, indicating good agreement between annotators. The figure \ref{fig:Dataset dist} depicts the distribution of data among multiple categories. All the disagreements between annotators were resolved by adjudication with the help of an expert. For details about the annotators, please refer \ref{sec:annotator_details}. 

Out of 5409, our dataset has \textbf{4325} bias instances ($~80\%$ of dataset) and \textbf{1084} neutral (not biased towards any identity). The number of instances for each category across train, dev., test are shown in Table \ref{tab:data dist}.

\begin{table}
\centering
\begin{tabular}{lllll}
\hline
Categories & train & dev & test & total\\ \hline
bias & 3460 & 346 & 519 & 4325  \\
neutral & 867 & 86 & 131 & 1084  \\ \hline
race & 1769 & 181 & 252 & 2202  \\
religion & 1257 & 120 & 196 & 1573  \\
gender & 293 & 24 & 41 & 358 \\
lgbtq & 453 & 41 & 82 & 576 \\ 
political & 172 & 20 & 26 & 218\\ \hline
\end{tabular}
    \caption{Distribution of different categories across three splits of Train, Dev. and Test. }
    \label{tab:data dist}
\end{table}

\subsection{Annotation Challenges}
Several issues were encountered while annotating harmful datasets, which must be handled to annotate and build a consistent dataset.
\begin{itemize}
\itemsep0em 
    \item Quoting someone else statements considered biased? Example:\\
\textit{Trump said "Mexicans are rapists and drug dealers".}

One may argue that such statements are not biased as they are merely quoting someone else's opinions that are not theirs. However, we label them biased because quoting someone else's statement is just as detrimental and harmful.

\item We believe that asking questions about an issue may not lead to bias. Example:\\
\textit{Black idiot or white idiot. What is the difference?
}

One would believe that this phrase is about idiots in general, without regard to race. As a result, there will be no prejudice. On the other hand, some may interpret that both blacks and whites are referred to as fools here.

\item We also encountered statements lacking context. Example:\\
\textit{Is that the white kind?  I mean since you hate whites so much?}

Now here, we do not know whether the statement is talking about white colour or white race. We label these kinds of sentences as neutral

\item Some statements were purely made as a personal attack. These instances were labeled appropriately as biased or neutral. Example:\\
\textit{Trump pig latin.  Oink, oink, oink, grab em by the poo say}

We label this statement as biased because Trump here represents a certain political party (community), unlike the below example:\\
\textit{settlers is a demeaning racist term. You Johnny are a white hating racist.}

Here Johnny is not a big political leader (name). So we mark this as neutral.

\item We have encountered many sarcastic instances in the dataset and label them appropriately. Example:\\
\textit{Ah yes, re-education! That's what us nasty white folks need.}

We label this statement as neutral because it's a self-criticism sarcastically.

\textit{Yeah - - because up until now, Islamic State really loved the US! And the West in general! They love us so much sometimes they cut off peoples heads to keep as a souvenir!
}

The above statement was labeled as biased as it is sarcastically showing prejudice against Islam.

\item Some statements are speaker dependent. Example:\\
\textit{Shit still happenin and no one is hearin about it, but niggas livin it everyday.}

This statement will not be biased if said by an African-American; however, it becomes highly offensive and biased if stated by someone else.

\end{itemize}

\section{Experimental Setup} \label{sec:expt}

\begin{table*}[!htp]

\centering
\begin{tabular}{llllll}
\hline
 & Model & P & R & F1 & Acc \\ \hline
 & Logistic Regression & 0.67 & 0.50 & 0.46 & 0.84 \\
 Baselines & SVM & 0.42 & 0.50 & 0.46 & 0.84 \\
 & Bi-LSTM + Glove& 0.59 & 0.58 & 0.58 & 0.78 \\ \cline{2-6}

 & BERT (Hierarchical) & 0.62 & 0.66 & \textbf{0.64} & \textbf{0.86} \\
Transformers w/o Aug & BERT (Multi-task) & 0.90 & 0.52 & 0.49 & 0.81 \\

 & GPT2 & 0.62 & 0.66 & 0.62 & 0.71 \\ \cline{2-6}
  & BERT (Hierarchical)& 0.86 & 0.86 & \textbf{0.86} & \textbf{0.88} \\
Transformers /w Aug  & BERT (Multi-task) & 0.86 & 0.86 & \textbf{0.86} & 0.87 \\

 & GPT2 & 0.81 & 0.86 & 0.84 & 0.81 \\ \hline
\end{tabular}
    \caption{Performance of various models on bias detection task. We report results for baselines, and Transformer based training. For Transformer based training, we compare performances without data augmentation and with data augmentation. Best scores are shown in bold.}
    \label{tab:Bias Detection Result}
\end{table*}

In this section we will discuss about different models trained for detection of social biases and their categories. For all our experiment, we split the data into train, development, and test (80:8:12) set. Since the dataset was imbalanced with respect to bias column, we split it in stratified manner.

\subsection{Metrices}
We report accuracy, macro F1-score, and AUC-based scores in accordance with best practice. These metrics would be used to assess the classifier's ability to distinguish between the bias and neutral texts along with bias categories. AUC stands for \textit{Area under the ROC curve.} ROC curve depicts the tradeoff between true positive rate (TPR) and false positive rate (FPR). The AUC value is high when the TPR is high and the FPR is low.

\citet{borkan} proposed AUC-based metrics to quantify the unintended model bias. These metrics compare the output distributions of instances that include the specific community word (subgroup distribution) with the rest (background distribution). The three AUC-based bias scores are as follows:
\begin{enumerate}
\itemsep0em 
  \item \textbf{Subgroup AUC ($AUC_{sub}$):} It calculates AUC exclusively on a subset of the data for a specified community word. A low score indicates that the model struggles to differentiate between bias and neutral comments related to the community word.
  \item \textbf{Background Positive and Subgroup Negative AUC ($AUC_{bpsn}$):} $AUC_{bpsn}$ uses the biased background instances and the neutral subgroup examples to determine AUC. A low score indicates that the model has high false positive rate. The model misinterprets neutral comments mentioning the community with biased comments missing it.
  \item \textbf{Background Negative and Subgroup Positive AUC ($AUC_{bnsp}$):} It uses the neutral background instances and the biased subgroup examples to determine AUC. A low score suggests that the model has a high rate of false negatives. The model misunderstands biased comments that mention the community with neutral ones that do not.
\end{enumerate}

\begin{table}[t]
\centering
\resizebox{\columnwidth}{!}{
    \begin{tabular}{lcccc|cccc}
    \toprule
    \multicolumn{1}{c}{\multirow{2}{*}{\textbf{Model}}} & \multicolumn{4}{c}{\textbf{Hierarchical}} & \multicolumn{4}{c}{\textbf{Multi-task}} \\ \cline{2-9}                
                \multicolumn{1}{c}{}  & \textbf{Acc}  & \textbf{P}    & \textbf{R}   & \textbf{F1}  & \textbf{Acc}  & \textbf{P}    & \textbf{R}   & \textbf{F1}  \\ \hline
                
     {political} & 0.96 & 0.48 &  0.50 & 0.49  & 0.96 & 0.77 &  0.57 & \textbf{0.61} \\                                                                                       
    {gender}   & 0.95  & 0.47  &  0.50 & 0.49 & 0.95  & 0.84  &  0.71 & \textbf{0.76} \\
                                                              
    {race}  & 0.84 & 0.81 & 0.83 & 0.82 & 0.86 & 0.86 & 0.88 & \textbf{0.86} \\ 
                                                                 
    {religion} & 0.82 & 0.82 & 0.82 & 0.82 & 0.93 & 0.91 & 0.94 & \textbf{0.92} \\
    
    {lgbtq} & 0.93 &  0.81 & 0.81 & 0.81 & 0.94 &  0.86 & 0.87 & \textbf{0.86}

                     \\ \bottomrule
    \end{tabular}
}
\caption{Bias Category Detection Results. P, R, F1 and Acc are Precision, Recall, F1-score and Accuracy respectively. Best scores are shown in bold.}
\label{tab:Bias Category Detection Results}
\end{table}

\subsection{Baseline Models}
We discuss several model architectures for detection of biases and their categories. For bias detection, which is a binary class classification task, we consider Logistic Regression (LR) with TF-IDF as our baseline model. Our baseline model gives 84\% accuracy with 0.46 F1 score. The low F1 score clearly indicates that model has very high false positive rate and false negative rate. We also tried Support Vector Machine (linear kernel) with TF-IDF and LSTM \cite{lstm} with Glove 300d word representation \cite{glove}. The best model is observed to be BERT \cite{devlin-etal-2019-bert} with 0.64 F1 score. Two different model settings were used to detect biases and their categories. We will discuss each of them in detail in the following sections.

\subsection{Hierarchical Model}
In the hierarchical model, bias detection and category classification was done in two steps. Bias detection, a binary class classification task, is performed first. If the post has some biases, then its categories are detected next. Since a post may contain several biases, the bias category detection task was framed as multi-label classification. Bias detection results of several models in hierarchical model architecture is shown in Table \ref{tab:Bias Detection Result}. Bias category detection results in the hierarchical setting are shown in Table \ref{tab:Bias Category Detection Results}.

\subsection{Multi-task Learning}
In the context of classification, multi-Task Learning tries to improve the performance of numerous classification problems by learning them together. So instead of predicting bias and its category in two steps, we can train a model to predict them simultaneously in one step. Since there can be multiple biases in a post, we cannot use logistic regression or SVM in a multi-label classification task. Hence in this model architecture, we try LSTM and BERT models only. We use LSTM with a single output layer. The last dense layer of the LSTM comprises six neurons, one to detect bias and the other five to identify bias categories.

Precision (P), recall (R), F1 (macro values for all), and accuracy (Acc) for bias detection experiments in Multi-task architecture is shown in Table \ref{tab:Bias Detection Result}. Table \ref{tab:Bias Category Detection Results} shows the comparison between hierarchical and multi-task model for category detection task. 

\begin{table*}[]
    \resizebox{\textwidth}{!}{%
    \begin{tabular}{ |l|c|c|c|c| }
    \hline
        comment\_text & \makecell[l]{Ground truth\\ label} & \makecell[l]{Predicted\\ label}& \makecell[l]{Ground truth\\ category} & \makecell[l]{Predicted\\ category}\\ \hline
        Quran is a holy book. & neutral & bias & - & religion \\
        \makecell[l]{ So then I was all like "I'd rather get the black plague \\ and lock myself in an iron maiden than go out with you.} & neutral & bias & - & race\\
        Do they come in men's sizes? & neutral& bias & - & gender\\
        \makecell[l]{What I've just shown is that this happens in every black hole.} & neutral & bias & - & race\\
    \hline
    \end{tabular}}
    \caption[Model Biases]{Error analysis showing model biases from predictions of Multi-task BERT model without augmentation.}
    \label{tab:Model Biases}
\end{table*}

\begin{table}
\centering
\begin{tabular}{llllll}
\hline
Variables & BLEU-2 & RougeL \\ \hline
Categories & 61.60$\pm$0.96 & 88.23$\pm$1.23 \\
Target subgroup & 52.95$\pm$2.84 & 77.58$\pm$4.21  \\
Implications & 33.4$\pm$1.55 & 39.5$\pm$1.20 \\ \hline
\end{tabular}
    \caption{Evaluation of various generation tasks. The standard deviations for three runs are also reported.}
    \label{tab:evaluation of generation}
\end{table}

\subsection{Generation Framework}
Considering the efficacy of GPT \cite{Radford2018ImprovingLU} based model for classification, conditional generation tasks \cite{sap2020social}, we frame the prediction of categorical variables and implications as generation task. The input is a sequence of tokens as in Equation\ref{gpt2-ip}, where $w_{i}$ are the tokens corresponding to comment text and $[\mathrm{BOS}], [\mathrm{SEP}], [\mathrm{EOS}]$ are start token, separator token and end token respectively. Two task specific tokens ($[\mathrm{BON}], [\mathrm{BOFF}]$) were added to the token vocabulary which were used as $w_{[\text {bias }]}$ in the input. Here, $[\mathrm{BON}], [\mathrm{BOFF}]$ correspond to bias and neutral instances respectively. As we have many inputs with multiple bias categories and targets, we combine them using a comma separator in the raw text. While encoding the input we use $w_{[\mathrm{C}]_{i}}, w_{[\mathrm{T}]_{i}}$ as the token corresponding to them respectively. Similarly, $w_{[\mathrm{R}]_{i}}$ is used for representing the tokens corresponding to implications.

\begin{equation} \label{gpt2-ip}
\begin{gathered}
\mathbf{x}=\left\{[\mathrm{BOS}], w_{i},[\mathrm{SEP}]\right. w_{[\text {bias }]}, [\mathrm{SEP}] \\
w_{[\mathrm{C}]_{i}},[\mathrm{SEP}] w_{[\mathrm{T}]_{i}},[\mathrm{SEP}] 
w_{[\mathrm{R}]_{i}},[\mathrm{EOS}] \}
\end{gathered}
\end{equation}

For this experiment, we finetune the GPT-2 \cite{gpt2} model with commonly used hyperparameters. For training we use cross-entropy loss as cost function. During inference, we first calculate the normalized probability of $w_{[\text {bias }]}$ conditioned on the initial part of input and then append the highest probable token to the input and generate rest of the tokens till $[\mathrm{EOS}]$.

We use BLEU-2 \cite{bleu} and RougeL (Fmeasure) \cite{rouge} as the metrics to calculate the performance of the model for category, target and implication of the comment text(Table \ref{tab:evaluation of generation}) and macro F1 as metric for bias evaluation(Table \ref{tab:Bias Detection Result}). Performance for category generation is better than other two variable as it has less ambiguity whereas the low performance for implications show the variability in the annotation for implications. 

We report all the hyperparameters and training setup in appendix \ref{sec:training_details}.

\subsection{Results and Discussion}

From the above tables, we can infer that the BERT-based hierarchical model outperforms Multi-task and GPT-2 models for the bias detection task. In addition, category identification is performed more effectively by multi-task model. This is quite apparent from the fact that in multi-task learning each model aid each others due to shared parameters. We also see that F1 scores for political and gender categories are less when compared to other categories in the category detection task. One plausible explanation might be the low frequency of such biases in our dataset. The BLEU scores of category and target subgroup generation are higher than those of bias implication generation, as shown in Table \ref{tab:evaluation of generation}. The fundamental reason for this is that the bias category and target have less distinct bi-grams/tri-grams than the implications. We show an error analysis for categorical prediction tasks after training the models. We discovered that the model predicts incorrectly for some simple sentences. A few examples are also listed in the Table \ref{tab:Model Biases}. This most likely occurred due to lexical overfitting/ model bias of the trained model towards some of the identity specific terminologies. In the next section, we discuss an approach to reduce this unwanted behavior of the model using counter-narrative augmentation.

\section{Mitigation of Model Bias} \label{sec:model_bias}

\begin{table*}[!htp]
    \resizebox{\textwidth}{!}{%
    \begin{tabular}{ |l|c|c|c| }
        \hline
        comment\_text & \makecell[l]{Ground truth\\ category} & \makecell[l]{Prediction (\textbf{without}\\ augmentation)} & \makecell[l]{Prediction (\textbf{with}\\ augmentation)}\\\hline
        Quran is a holy book. & none & religion & none \\
        Islam is a peaceful religion. & none & religion & none \\
        Muslims are invaders. & religion & religion & religion\\
        Do they come in men's sizes? & none & gender & none \\
        What I've just shown is that this happens in every black hole. & none & race & none \\
        \makecell[l]{I'm sick of niggers, it's a bad neighborhood. I don't like being\\ around them, they're animals.} & race & race & race \\

    \hline
    \end{tabular}}
    \caption[Model Bias Mitigation]{Error analysis showing the effectiveness of counter-narrative augmentation. After training with augmented data, the resulting model does not latch on to the community related terms.}
    \label{tab:Model Bias Mitigation}
\end{table*}

\begin{table}[!htp]
    \centering
    \begin{tabular}{|l|c|c|}
        \hline
        Subgroup & Count & \% Bias \\ \hline
        all comments & 5409 & 79.92 \\
        black & 1291 & 79.55 \\
        jewish & 269  & 74.34 \\
        lgbt & 778 & 77.24 \\
        muslim & 1263 & 87.01  \\
        female & 586 &  76.45 \\
        \hline
    \end{tabular}
    \caption[Percentage of bias comments]{Percentage of bias comments by identity terms such as black, jewish, lgbt, muslim, female in the \textit{ToxicBias} dataset.}
    \label{tab:Percentage of bias comments}
\end{table}

When we look at the incorrectly classified comments in Table \ref{tab:Model Biases}, we observe that they contain community words such as 'blacks,' 'Quran,' and so on. Sometimes, due to the presence of these community terms, our model predicts that these comments would be biased. In essence, our initial model is latching onto some community related terms and hence suffers from model bias. According to \cite{zueva-etal-2020-reducing}, most existing models provide predictions with certain bias. Even if the statement itself is not toxic, the model commonly classifies it as toxic if it includes specific frequently targeted identities (such as women, blacks, or Jews). Similarly, our model incorrectly labels comments referencing particular identities, such as Blacks, Muslims, and Whites, as social bias. Model biases emerge when identity words like Blacks, Whites, and Muslims appear more frequently in biased comments than in neutral comments. If the training data for a machine learning model is skewed towards certain terms, the final model is likely to acquire this bias. Table \ref{tab:Percentage of bias comments} shows the bias percentage in \textbf{\textit{ToxicBias}} for several identities/subgroups, indicating the imbalance for bias labels among those identities and emphasising the importance of AUC-based metrics resilient to these data skews. 

\noindent \textbf{Counter-narratives:} Despite enormous attempts to build suitable legal and regulatory responses to hate content on social media platforms, dealing with hatred online remains challenging. If hate speech is addressed with standard content deletion or user suspension methods, censorship may be accused. Actively addressing hate material through counter-narratives (i.e., informed textual responses) is one potential technique that has received little attention in the academic community thus far. A counter-narrative (also known as a counter-comment or counter-speech) is a reply that provides non-negative feedback through fact-based arguments and is often recognized as the most effective way to deal with hate speech.

\begin{table}[!htp]
    \centering
    \resizebox{\columnwidth}{!}{%
    \begin{tabular}{lcccc}
        \hline
        subgroup & $AUC_{sub}\uparrow$ & $AUC_{bpsn}\uparrow$ & $AUC_{bnsp}\uparrow$\\ \hline
        black & 0.48 & 0.50 &  0.49  \\
        jewish & 0.47  & 0.50  &  0.49 \\
        lgbt & 0.81 & 0.83 & 0.82 \\
        muslim & 0.82 & 0.82 & 0.82 \\
        female & 0.81 &  0.81 & 0.81 \\
        \hline
    \end{tabular}}
    \caption[Model Bias without CONAN]{AUC based scores for subgroups on bias detection model trained without data augmentation.  Higher AUC values for
each target subgroup indicate reduced lexical overfitting/model bias for those targets.}
    \label{tab:Model Bias without CONAN}
\end{table}

\begin{table}[!htp]
\centering
    \resizebox{\columnwidth}{!}{%
    \begin{tabular}{lcccc}
        \hline
        subgroup & $AUC_{sub}\uparrow$ & $AUC_{bpsn}\uparrow$ & $AUC_{bnsp}\uparrow$\\ \hline
        black & 0.86 & 0.78 &  0.97  \\
        jewish &  0.91 & 0.93 & 0.91 \\
        lgbt & 0.89  & 0.91  &  0.93 \\
        muslim & 0.96 & 0.97 & 0.86 \\
        female & 0.93 &  0.94 & 0.93 \\
        \hline
    \end{tabular}}
    \caption[Model Bias with CONAN]{AUC based scores on bias detection model trained after data augmentation. Higher AUC values for each target subgroup indicate reduced lexical overfitting/ model bias for those targets.}
    \label{tab:Model Bias with CONAN}
\end{table}

We use two counter-narrative datasets to reduce the model biases: CONAN \cite{chung-etal-2019-conan} and Multi-target CONAN \cite{fanton-2021-human}. These datasets provide counter-narratives to hate speech or stereotypes directed towards social groups such as Muslims, Blacks, Women, Jews, and LGBT people. So they do not contain any negative social biases towards those groups. Combining these counter narratives ensures that the resulting dataset will have more neutral/positive instances mentioning those identity terms. Adding these counter narratives to our dataset significantly decreased model biases. We used total of $7219$ counter-narratives related to jews ($593$), muslim ($4996$), black ($352$), homosexual\_gay\_or\_lesbian ($617$), and female ($661$). As illustrated in table \ref{tab:Model Bias without CONAN}, black and jewish identities suffer from both high false positives and high false negatives. However, after counter-narrative augmentation, the resulting model appears to be capable of dealing with the problem of model bias. Table \ref{tab:Model Bias with CONAN} shows the reduction in model bias using AUC-based metrics. Table \ref{tab:Model Bias Mitigation} includes an error analysis to show how CONAN has helped reduce model bias.

\section{Conclusion and Future Work} \label{sec:con}
We have demonstrated that identity attacks or hate speech often incorporate social biases or stereotypes. However, not all hate speech can be labeled as social bias. Some of them are merely personal insults. Filtering out such biases from hate speech is not a trivial task. Furthermore, we have frequently observed that detecting bias without context for the comment or demographic information of the comment holder makes the annotation much more challenging. However, detecting these social biases from toxic datasets, which are available in relatively large amounts, will be a useful starting point for social bias research in other forms of text.

 The issue of model bias is also observed during inference. The imbalanced existence of particular community terms (muslims, whites, etc.) might lead to a model labeling a comment as biased. To attenuate model biases, we used counter-narratives and showed that they help significantly to reduce model biases. From our study, we also observe that biases can have directions too. So basically, biases can occur against specific communities and in favour of a community. We intend to detect such biases in future work.

\section{Acknowledgements}

We would like to the the anonymous reviewers as well as the CoNLL action editors. Their insightful comments helped us in improving the current version of the paper. Additionally, we would like to thank Sandeep Singamsetty, Prapti Roy, Sandhya Singh for their contributions in data annotation and useful comments. This research work was supported by Accenture Labs, India.

\section{Limitations} \label{sec:limitation}
The most notable limitation of our work is the lack of external context and small-sized dataset. In our present models, we have not considered any external context that can be useful for the categorization task, such as the profile bio, user gender, post history, etc. Our work currently considers only five types of social biases, not all other possible dimensions of bias. We also concentrated on using only the English language in our work, and the dataset is oriented toward western culture. The bias annotations in the dataset may not be very relevant to people of non-western culture. Furthermore, Multilingual bias is not taken into account.

\bibliography{custom}
\bibliographystyle{acl_natbib}

\appendix
\section{Appendix}\label{sec:appendix}

\subsection{Ethical Considerations}
Our work aims at capturing various social biases in toxic social media posts and demonstrates the annotation quality on biases in one of existing dataset. We also discuss the challenges we faced while doing the annotation of the dataset, specifically due to the absence of context for each instance in the dataset. Also, study of social biases come with ethical concerns of risks in deployment \cite{Ullmann2020}. As these toxic posts can create potentially harm to any user or community, it is required to conduct this kind of research to detect them. If done with precautions, such research can be quite helpful in automatic flagging of toxic and harmful online contents.

Researchers working the problem of social bias detection on any form of text would benefit from the dataset we have collated and from the inferences we got from multiple training strategies.

\subsection{Annotator Demographics and Treatment} \label{sec:annotator_details}
Both the annotators were trained and selected through extensive one-on-one discussions, and were working voluntarily. Both of them went through few days of initial training where they would annotate many examples which would then be validated by an expert and were communicated properly about any wrong annotations during training. As there are potential negative side effects of annotating such toxic comments, we used to have regular discussion sessions with them to make sure they are not excessively exposed to the harmful contents. Both the annotators were Asian male and were of age between $23$ to $26$. The expert was an Asian female with post-graduation degree in sociology.

\subsection{Training Details} \label{sec:training_details}
\subsubsection{BERT Training}
We finetune $12$ layer BERT base uncased with batch size of $32$ for two epochs. Max token length of $128$ is used. We experiment with learning rates of ${2e-5,3e-5, 4e-5, 5e-5}$ with AdamW\cite{adamw} optimizer and epochs of ${5, 10, 20}$. We also use a dropout layer in our model. AdamW optimizer with learning rate = $5e-05$, epsilon = $1e-08$, decay = $0.01$, clipnorm = $1.0$ were used.
\subsubsection{GPT-2 Training}
We finetune GPT-2 with a training batch size of $1$, gradient accumulation step as $4$, and $200$ warm up steps. Experiments were run with a single GeForce RTX 2080 Ti GPU. Finetuning one GPT-2 model took around $40$ minutes for $5$ epochs. 

We have kept all the parameters of BERT and GPT-2 trainable. All of our implementations uses Huggingface's transformer library \cite{wolf2020huggingfaces}.

\end{document}